# Reinforcement Learning Platform for Adversarial Black-box Attacks with Custom Distortion Filters

**Soumyendu Sarkar**[1*†], **Ashwin Ramesh Babu**[1†], **Sajad Mousavi**[1†], **Vineet Gundecha**[1], **Sahand Ghorbanpour**[1], **Avisek Naug**[1], **Ricardo Luna Gutiérrez**[1], **Antonio Guillen**[1], **Desik Rengarajan**[12]

[1]Hewlett Packard Enterprise
[2]Amazon
{soumyendu.sarkar, ashwin.ramesh-babu, sajad.mousavi, vineet.gundecha, sahand.ghorbanpour,
avisek.naug, rluna, antonio.guillen, desik.rengarajan}@hpe.com

## Abstract

We present a Reinforcement Learning Platform for Adversarial Black-box untargeted and targeted attacks, RLAB, that allows users to select from various distortion filters to create adversarial examples. The platform uses a Reinforcement Learning agent to add minimum distortion to input images while still causing misclassification by the target model. The agent uses a novel dual-action method to explore the input image at each step to identify sensitive regions for adding distortions while removing noises that have less impact on the target model. This dual action leads to faster and more efficient convergence of the attack. The platform can also be used to measure the robustness of image classification models against specific distortion types. Also, retraining the model with adversarial samples significantly improved robustness when evaluated on benchmark datasets. The proposed platform outperforms state-of-the-art methods in terms of the average number of queries required to cause misclassification. This advances trustworthiness with a positive social impact.

## Introduction

Despite deep learning models demonstrating impressive performance in various tasks, they remain highly susceptible to input data corruption. This susceptibility is particularly concerning in safety-critical applications such as self-driving cars, facial recognition systems, and image-based authentication. The possibility of natural and domain-specific distortions at deployment poses a significant challenge for these models. Therefore, measuring the robustness of deep learning models against distortions becomes essential to uncover vulnerabilities and limitations of poorly trained models.

In the field of adversarial attacks, white-box attacks have been widely explored (Szegedy et al. 2013)(Goodfellow, Shlens, and Szegedy 2014) (Chakraborty et al. 2018) (Su, Vargas, and Sakurai 2019). These attacks require complete knowledge of the target model, including its architecture and parameters, to assess its vulnerability against distortions. However, access to such detailed information about the model is often restricted in real-world scenarios due to intellectual property concerns and support issues. This limitation makes white-box attacks less practical for evaluating models in many applications.

---
[*]Corresponding Author
[†]These authors contributed equally.

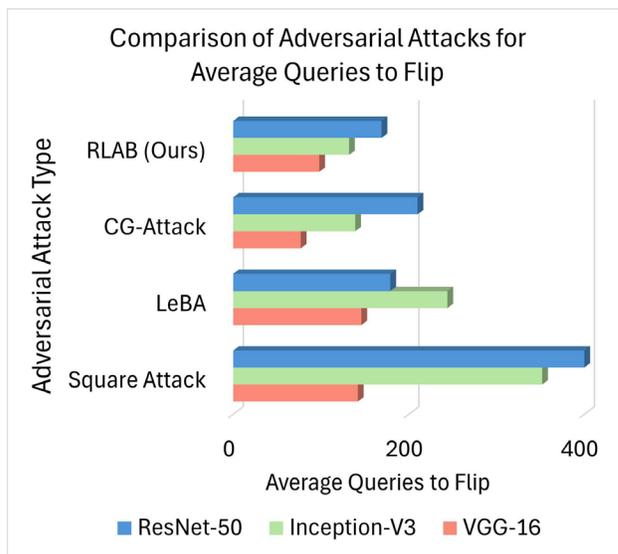

Figure 1: RLAB achieves the lowest average number of queries among untargeted black-box $L_2$-attacks on ImageNet across three CNN models with a Gaussian noise filter.

In contrast, black-box attacks do not require complete visibility into the target model. These attacks operate with limited information and exploit vulnerabilities by interacting with the model through input queries. However, black-box attacks suffer from inefficiency, often necessitating a large number of queries to create adversarial samples that can effectively compromise the evaluated model. Furthermore, many state-of-the-art black-box attack approaches rely on specific unnatural distortions based on hand-crafted heuristics, limiting their applicability and effectiveness.

In this work, we propose a Reinforcement Learning agent for a Platform (RLAB) that can learn a policy to make an adversarial attack with fewer queries and a high success rate. This RL agent is versatile enough to allow the user to plug in one or more of their own distortion filters in the spirit of "Bring Your Own Filter" (BYOF). It optimizes policy and action to adapt to many noise filters, generating efficient adversarial samples. The platform includes a dual-action RL Agent, which makes parallel addition and removal of distortions based on the image region sensitivity at each step and the history of the progression of added distortions. The goal is to cause a misclassification with a minimum number of queries and a high success rate while constraining the difference between the original and adversarial samples to a minimum calculated using the $L_p$ norm. In an extensive evaluation with ImageNet and CIFAR-10 datasets on CNN architectures such as ResNet-50, Inception-V3, and VGG-16, RLAB outperforms state-of-the-art methods for the $L_2$ threat model on the number of queries and success rate while achieving competitive $L_p$ norm as shown in Figure 1.

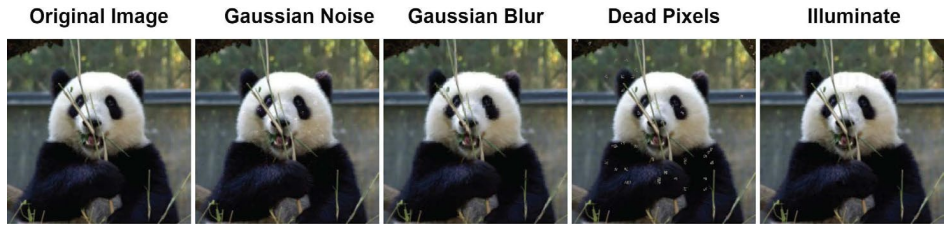

Figure 2: Example of adversarial samples generated by RLAB's RL agent for different types of filters.
**Demo:** https://tinyurl.com/55xxzceb

The main contributions of the work can be summarized as follows.

1. A novel Reinforcement Learning agent for a platform that can accept many types of distortion filters and can perform an efficient un-targeted or targetted black-box adversarial attack with optimized performance.

2. A dual action RL algorithm with a learned policy for an adversarial attack agent that can outperform the prevailing state-of-the-art black-box adversarial attacks by metrics such as average query and average success rate, with a competitive $L_p$ norm.

## Related Works

Traditional metrics like accuracy, precision, recall, and F1 score often fail to capture vulnerabilities exposed by adversarial examples. Szegedy et al. (Szegedy et al. 2013) first demonstrated adversarial attacks, and Goodfellow et al. (Goodfellow, Shlens, and Szegedy 2014) introduced the widely-used Fast Gradient Sign Method (FGSM) for white-box attacks.

Building on this work, subsequent studies explored gradient-based distortions to mislead models (Kurakin et al. 2016; Kurakin, Goodfellow, and Bengio 2016; Dong et al. 2018). Papernot et al. (Papernot et al. 2016) introduced a saliency map to identify vulnerable regions of the input for targeted attacks. Similarly, Moosavi et al. (Moosavi-Dezfooli, Fawzi, and Frossard 2016) proposed DeepFool, a straightforward yet effective method for adding perturbations to deceive machine learning models.

Black-box attacks operate with limited or no visibility into the model. In partially visible cases, information like loss functions, prediction probabilities, or top-K labels may guide query-based attacks. Comprehensive surveys by Michel et al. (Michel, Jha, and Ewetz 2022) and Chakraborty et al. (Chakraborty et al. 2018) highlight trends in adversarial attacks, while Ilyas et al. (Ilyas et al. 2018) tackled constraints like limited visibility and query access.

Notable black-box methods, including Square Attack (Andriushchenko et al. 2020), SimBA (Guo et al. 2019), and LeBA (Yang et al. 2020), operate within fixed $L_2/L_\infty$ budgets and successfully target convolutional networks. Guo et al. (Guo et al. 2019) iteratively sampled vectors from an orthonormal basis to modify images, while Andriushchenko et al. (Andriushchenko et al. 2020) used square-shaped updates at random positions under budget constraints.

Recent advancements, such as EigenBA (Zhou et al. 2022), Pixle (Pomponi, Scardapane, and Uncini 2022), QueryNet (Chen et al. 2021), AdvFlow (Mohaghegh et al. 2020), and CG Attack (Feng et al. 2022), achieve state-of-the-art results. Most methods rely on unnatural distortions, which may not generalize across use cases (Ratner et al. 2017; Shijie et al. 2017), emphasizing the need for adaptable platforms. Natural perturbations like Gaussian noise, blur, or brightness changes could offer more practical alternatives.

Most state-of-the-art approaches focus on adding similar but unnatural distortions to the input to generate adversarial samples. There is no guarantee that they would still work if it is applied for a different use case (Ratner et al. 2017)(Shijie et al. 2017). This raises the need for a common platform that allows switching the type of distortion used based on the actual needs of individual use cases. Also, adversarial attacks using naturally occurring perturbations such as Gaussian noise, blur, changes in brightness, and dead pixels may be more useful.

### Reinforcement Learning for Adversarial Attacks

Reinforcement Learning (RL) has demonstrated success in solving problems where classical machine learning often falls short, with applications spanning robotics, healthcare, controls, energy, and medical imaging. However, its potential in adversarial attacks remains underexplored. Sun et al. (Sun et al. 2020) applied RL to target graph neural networks through node injection, while Yang et al. (Yang et al. 2020) (Patch Attack) utilized RL to attack CNN models by overlaying textured patches on input images, though their method struggled to minimize distortions.

In contrast, our RL agent employs a detailed state representation that captures the model's sensitivity to different image regions and facilitates a patch-based attack process with any distortion type. This approach significantly surpasses state-of-the-art adversarial attacks, even when using naturally occurring distortions.

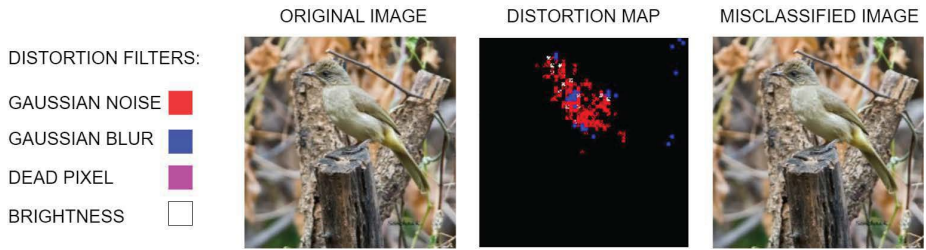

Figure 3: Mix of distortion filters for Adversarial Attack. Steps: 176, L2: 4.45. **Demo:** https://tinyurl.com/24ww544s

## RLAB Platform

### Problem Formulation

A Deep Neural Network (DNN) model under evaluation can be expressed as $y = argmax f(x; \theta)$, where $x$ represents the input image, $y$ represents the prediction, and $\theta$ represents the model parameters. A non-targeted black-box attack without access to the $\theta$ generates a perturbation $\delta$ such that, $y' = f(x + \delta; \theta)$. The distance between the original and adversarial sample, $D(x, x + \delta)$ will be any function of the $l_p$ norms. The objective is to fool the classifier while keeping $D$ to a minimum.

### Bring Your Own Filters (BYOF)

The RLAB platform is highly versatile because the user can use it with any type of distortion of their choice. The RL algorithm learns a policy to adapt to the filter used such that the adversarial samples are generated with minimum distortion $D$. Further, the algorithm can be used with a mixture of filters such that the agent first decides on which filter (Gaussian Noise, Gaussian Blur, brightness, etc.) to use for every step and further decides on the number of patches to which the filter needs to be added. We experimented with multiple filters and have presented four naturally occurring distortion filters as part of this paper. Figure 2 shows adversarial examples generated using different filters. Figure 3 shows adversarial examples generated with a mix of different distortion filters.

### How it works

In RLAB, the input image is divided into square patches of size $n \times n$ and the sensitivity of the ground truth probability $P_{GT}$, to the addition and removal of various types of distortions is estimated for the patches. Using the sensitivity information, the RL agent performs two actions,

- Adds distortions to selected patches.
- Removes distortions from selected patches.

This process is repeated iteratively until the model misclassifies the image or the budget for the maximum allowed steps is exhausted. Once an adversarial sample is successfully generated (i.e., the model misclassifies), we apply an iterative image cleanup as a post-processing step to further minimize $D$. In the mixed filter setting, the RL agent also selects the optimal distortion filter for each step. The overall flow of the proposed method is illustrated in Figures 4 and 5.

## RL for Adversarial Attack

### State Design

We designed a state space that balances sufficient visibility for the RL agent with simplicity for efficient training. Sensitivity analysis is used to identify key patches in the input image.

**Sensitivity Analysis:** Distortion filters (masks) of size $n \times n$ matching the square patches, are generated for sensitivity analysis. Each filter has fixed hyperparameters, such as noise or brightness levels, throughout the experiment. During training and validation, the mask is applied across all patches to measure the change in the ground truth classification probability $P_{GT}$. The hyperparameters associated with the distortion filters (noise intensity, amount of blurring, etc.) are kept minimal to allow finer-grained distortion additions in successive steps, aiding in controlling the $L_p$ norm. The distorted samples are constrained to the values $[0, 1]^d$. In a multi-filter setting, where the RL agent has a choice of filters, the hyperparameters for the individual filters were chosen such that the impact on $L_p$ norm is the same after adding any filter. Additional experiments were performed for different types of patch sizes as represented in the table 5 to compare the effect of the patch sizes.

**State Vector:** The state vector incorporates the results of the image sensitivity analysis, ordered based on the drift in $P_{GT}$ for patches during addition ($LIST_{ADD}$) and removal ($LIST_{REMOVE}$) of distortions. It also includes the classification probabilities for each class at each step ($LIST_{PROB}$) and the $L_p$ norm, as illustrated in Figure 6.

**Inexpensive Query on a GPU:** The sensitivity analysis takes just a few GPU cycles on a V100 GPU, as the operation is a fully parallelized one-shot filter operation on an image. Also, multiple images can be processed in parallel on a V100 32GB GPU based on image size. So, processing a query to get the states is a fast operation on a GPU. This makes the execution very fast and efficient.

### Action

At each step, the RL agent selects the number of patches ($N_{ADD\_DIST}$) from $LIST_{ADD}$ to which distortion will be added, and the number of patches ($N_{REM\_DIST}$) from $LIST_{REMOVE}$ from which distortions will be removed, as shown in Figure 7. The RL action space was designed to be discrete and simple, ensuring the RL policy is learnable. By design, $N_{REM\_DIST} < N_{ADD\_DIST}$, ensuring that distortions are progressively added at each step to minimize the number of queries. However, distortions may be removed from patches that previously had multiple distortions added, thereby reducing the net increase $L_2$ distance for the step. To keep the computation bounded, the action space is constrained to $N_{ADD\_DIST} \in [1, N_{max}]$, where $N_{max}$ is a hyperparameter and is set to 8 for ImageNet (224 × 224) image size with 2×2 patch size), to balance effectiveness and computation.

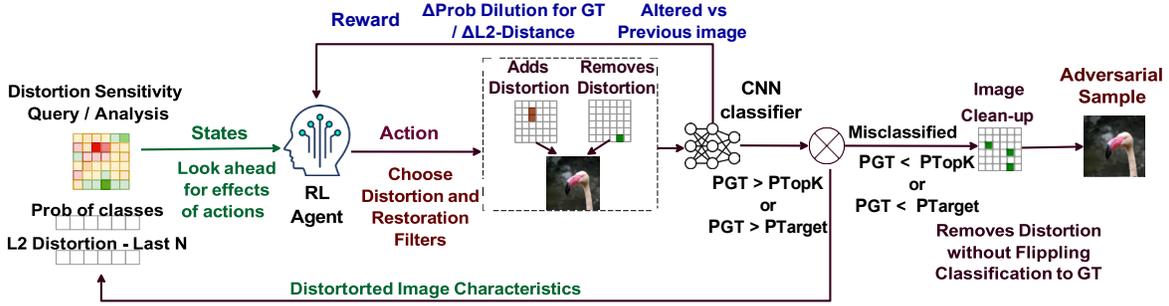

Figure 4: Workflow with Reinforcement Learning overview for RLAB.

In a mixed filter setting, the action is divided into two stages: first, the RL agent selects the distortion filter type for the step, and then distortions are added and removed as described above.

**Alternate to Tree Search** The intuition behind having two actions (addition and removal) is inspired by the application of reinforcement learning for board games. For board games, the most effective moves or actions are figured out through a Deep Tree Search (DTS) of multiple layers at the current step on a longer time horizon as the game evolves. DTS is computationally expensive, even with approximations like Monte Carlo Tree Search (MCTS). But unlike a board game, in this problem, there is a possibility to reset the earlier moves when we realize that we have made a less optimized move a few steps back. In the RLAB platform, this is done by removing distortions from patches to which distortions were added in the earlier step and adding distortions to other patches, considering the state of the modified image at any given step (equivalent to position on the board). This is equivalent to replaying all the moves in one step while keeping the sensitivity analysis restricted to the current state of the image without a tree search.

Our method reduces the complexity from $O(N^d)$ to $O(N)$ where N represents the computation complexity of one level of evaluation and corresponds to the image size, and d represents the depth of the tree search, which translates to how many queries and actions we would like to look ahead if we were doing a tree search. d=[1, max_steps].

**Reward**

We define a Probability Dilution (PD) metric to quantify the shift in classification probability from the ground truth to other classes (untargetted) or, to a specific class (targetted). The change in PD resulting from an action, denoted as ($\Delta$PD), measures the effectiveness of that action. Additionally, the cost of an action is defined as the change in $L_2$-distance ($\Delta L_2$), which represents the distortion introduced. The reward is then calculated as the normalized PD, as shown in equation 2.

$$PD_{untargeted} = -\frac{1}{\log \frac{1}{p_g}} + \frac{1}{\log \frac{1}{p_k}} + \frac{1}{\log \frac{1}{p_k}} + \ldots + \frac{1}{\log \frac{1}{p_{kn}}} \quad (1)$$

where, $p_g$ is the probability score of the ground truth class, $p_{k1}$ is the probability score of the second best class and $p_{kn}$ is the probability score of the $n^{th}$ class where $n$ is a hyperparameter.

$$R_t = \Delta PD_{normalized} = \Delta PD / \Delta L_2 \quad (2)$$

The change in the distribution of the probabilities across classes is updated in the state vector($LIST_{prob}$) at every step

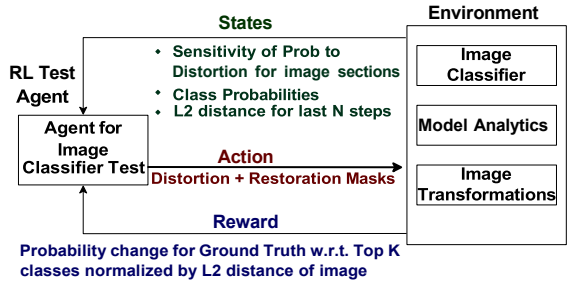

Figure 5: RL states, action, and rewards RLAB

| | |
|---|---|
| $LIST_{ADD}$ | Square patches in descending order of normalized sensitivity to addition of distortion |
| $LIST_{REMOVE}$ | Square patches in ascending order of normalized sensitivity to removal of distortion |
| $LIST_{L2}$ | L2 distance from original for the last Nsteps = 4 steps |
| $LIST_{PROB}$ | Classification probability of various classes at this step |

Figure 6: Reinforcement Learning states for RLAB

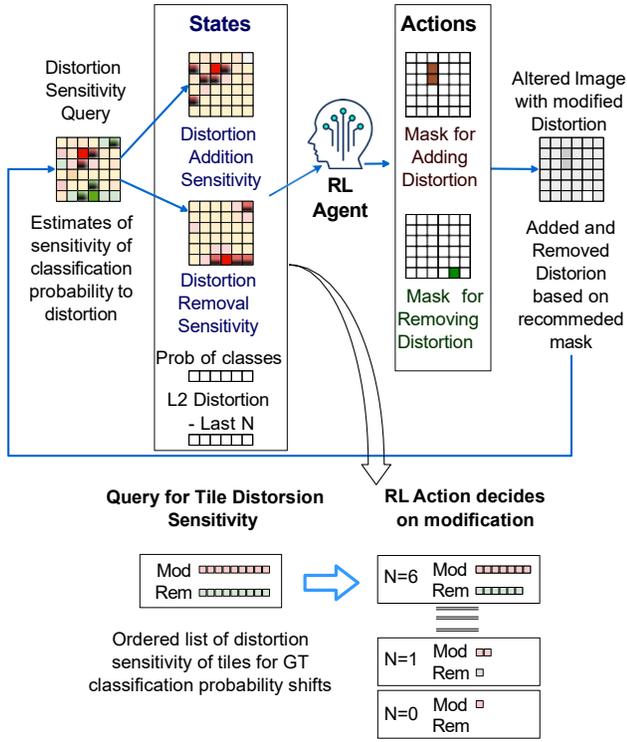

Figure 7: Dual Action for Reinforcement Learning agent: addition and removal of distortion

Algorithm 1: Reinforcement Learning Training in RLAB
**Initialization:** Policy parameters
**Input:** Validation set, number of iterations $Max_{iter} = 3500$
**Output:** Optimized policy for Dueling DQN

1: **for** image in validation set **do**
2:     Load the image
3:     Calculate reward $R_t$ and advantage $\hat{A}_t$ based on current value function
4:     Calculate sensitivity of ground truth classification probability $P_{GT}$ to change in distortion for square patches **for various distortion types**
5:     $i \leftarrow 0$
6:     $Pred_{fstep} \leftarrow 1 - P_{GT}$
7:     **while** $Pred_{GT} == Pred_{fstep}$ **and** $i < Max_{iter}$ **do**
8:         Collect set of trajectories (state, action) by running policy $\pi_k = \pi(\theta_k)$ in the environment → action $(N_{add\_dist}, N_{rem\_dist})$
9:         Calculate reward $R_t$ and TD error
10:        Update the DQN policy
11:        Compute/take action and perform prediction $Pred_{fstep}$
12:        $i \leftarrow i + 1$
13:     **end while**
14: **end for**

such that the RL agent can choose the optimum action at every step, maintaining the $L_p$ and the number of step/queries. Similarly, for a targeted attack, the reward can be defined as,

$$PD_{targeted} = -\frac{1}{\log\frac{1}{p_{tgt}}} + \frac{1}{\log\frac{1}{p_k}} + \frac{1}{\log\frac{1}{p_k}} + ..... + \frac{1}{\log\frac{1}{p_{kn}}} \quad (3)$$

$$R_t = \Delta PD_{normalized} = \Delta PD/\Delta L_2 \quad (4)$$

where $p_{tgt}$ represents the probability score of the target class. The goal here is to choose the patch to which, when noise is added, there is the highest increase in the target class. Through hyperparameter tuning, we obtained a discount factor $\gamma = 0.95$, where $\gamma$ determines how much the RL agent cares about rewards in the distant future relative to those at the current step.

## RL Algorithm

We utilized a Dueling DQN algorithm-based Reinforcement Learning (RL) agent for RLAB as the adversarial attack agent (Sewak 2019), which also evaluates the robustness of CNN image classification models. The Dueling DQN model is well-suited to the discrete action space, accommodating the limited possible values of $N_{ADD\_DIST}$ and $N_{REM\_DIST}$, while maintaining an appropriate level of complexity for effective prediction within a reasonably bounded training process. The overall training procedure for the proposed approach is detailed in Algorithm 1.

## Experimental Details

In this section, we discuss the effectiveness of our proposed method with the same experimental setup as our competitors. We evaluate two popular image classification datasets, ILSVRC2012 (Russakovsky et al. 2015) and CIFAR-10. Eighty percent of the original validation set was used to train the RL algorithm, and 20 percent was used for evaluation. We performed our attacks on three major Convolution-based Neural Network architectures: ResNet-50, Inception-V3, and VGG-16 for both targeted and un-targeted attacks. We used **three metrics** to evaluate the performance of our approach, $L_p$ norm, which is a measure of distortion, the average number of queries (AVG.Q) to make a model missclassify a correctly classified sample, and the average success rate (ASR).

With the values of pixels set to a range between 0 and 1 and a maximum query budget of 3500, 1000 samples from the ImageNet dataset on ResNet-50 architecture are evaluated. A failure case is when the proposed method could not fool the victim model within the given budget, and failure cases were not included in any of the metrics calculated except for the success rate. All experiments were performed for a patch size of $2 \times 2$ and a Gaussian noise-based distortion filter, as we got the best results for this configuration.

The pipeline computation is executed on GPU and efficiently parallelized, batched, and scaled on GPUs. Caching techniques, such as noise masks, were used for pre-computed information for improved efficiency. Apollo servers with $8 \times V100$ 32 GB GPUs were used for training and validation. We processed 16(images per GPU) x 8(GPUs) = 128 images in a batch for the complete pipeline.

It is worth mentioning that the proposed robustness measure in DeepFool [4] involves minimizing the amount of distortion needed for misclassification, which is defined by $\Delta(x, \hat{k}) := \min_r \|r\|_2$ subject to $\hat{k}(x + r) \neq \hat{k}(x)$, where $\min_r \|r\|_2 = \min D$ and $\Delta(x, \hat{k})$ is the robustness of classifier $\hat{k}$ for input x. As we can see, this is consistent with our goal, which is minimizing $D$.

## Results and Discussion

### Evaluation on ImageNet

| Untargeted Attack | Avg.Q | $L_2$(avg) | $L_{inf}$ | ASR |
|---|---|---|---|---|
| Q-Fool (Chen, et al. 2020) | 5000 | 7.52 | - | - |
| NES (Ilyas et al. 2018) | 1632 | - | 0.05 | 82.7 |
| Bandits (Ilyas, et al. 2018) | 5251 | 5 | - | 80.5 |
| Subspace (Guo, et al. 2019) | 1078 | - | 0.05 | 94.4 |
| P-RGF$_D$ (Cheng et al. 2019) | 270.5 | 16.43 | - | 99.3 |
| TIMI (Dong et al. 2019) | 68.6 | - | - | - |
| LeBA (Yang et al. 2020) | 178.7 | 16.37 | - | 99.9 |
| Square attack (Andriushchenko et al. 2020) | 401 | 5 | 0.05 | 99.8 |
| SimBA-DCT (Guo et al. 2019) | 1665 | 5(3.98) | - | 98.6 |
| querynet (Chen et al. 2021) | - | 5 | - | - |
| (Mohaghegh, et al. 2020) | 746 | - | - | 96.7 |
| EigenBA (Zhou et al. 2022) | 518 | 5(3.6) | 0.05 | 98 |
| Pixle (Pomponi, et al. 2022) | 341 | - | - | 98 |
| CG-Attack(Feng et al. 2022) | 210 | - | - | 97.3 |
| Patch Attack (Yang et al. 2020) | 983 | - | - | - |
| **RLAB (Gaussian Noise)** | **178** | **4.74 (2.48)** | **0.07** | **100%** |

Table 1: Comparison of maximum $L_2$, $L_{inf}$, average queries and average success rate of the proposed method with state-of-the-art on the **ResNet-50** model trained on **ImageNet** dataset. L2 represents maximum L2 over all samples, (avg) represents average L2 over all samples. Results for Inception-V3 and VGG-16 have been included in the supplemental material.

Table 1 and 2 aggregate the proposed method's results compared to other state-of-the-art black-box algorithms on the ImageNet dataset for ResNet-50 and Inception-V3 architectures. Results for other architectures are included in the supplemental document. The competitors' results were generated with the best parameters described in their papers. The Average Success Rate (ASR) and Average Query (AVG.Q) were calculated for each victim model, while the maximum L2/$L_{inf}$ for most of the competitors were presented in their paper. It can be observed that our proposed approach beats state-of-the-art algorithms for average queries and success rates by a significant margin. Also, for the proposed approach, the maximum $L_2$ was 4.74, which is below the maximum budget (i.e., 5) of the state-of-the-art approaches. It is also worth mentioning that the proposed approach was able to achieve a 100% success rate for a maximum query set to 3500, while the competitors performed experiments with a maximum query set to 10000.

As the sensitivity analysis takes just a few GPU cycles on a V100 GPU, as the operation is a fully parallelized one-shot filter operation on an image, the queries are very fast, inexpensive, and efficient on a GPU. In addition, multiple images can be processed in parallel on a V100 32GB GPU based on the image size.

Our max value for $L_{inf}$ is marginally higher than the competitor's maximum budget for the ImageNet dataset. This could be because of the way $L_{inf}$ is computed (returns the maximum change from the original image). Perturbation that can affect the entire image by a very small value will have a smaller $L_{inf}$ value but a higher $L_2$. Our approach focuses on exploring and attacking only vulnerable regions in the input image as represented in Figure 3 leading to marginally higher $L_{inf}$.

| Targeted Attack | Avg.Q | ASR |
|---|---|---|
| NES (Ilyas et al. 2018) | 4944 | 79.4 |
| MetaAttack (Du et al. 2019) | 8341 | 17.5 |
| SignHunter (Al-Dujaili and O'Reilly 2020) | 1115 | 100 |
| Square attack (Andriushchenko et al. 2020) | 1112 | 100 |
| SimBA-DCT (Guo et al. 2019) | 6569 | 71.7 |
| CG-Attack(Feng et al. 2022) | 2447 | 92.9 |
| **RLAB (Gaussian Noise)** | **180** | **100%** |

Table 2: Comparison of average queries and average success rate of the proposed method with state-of-the-art on the **Inception-V3** model trained on **ImageNet** dataset for **Targeted Attack**. Results for other models have been included in the supplemental material.

### Evaluation on CIFAR-10

| Attack | AVG. Q | ASR |
|---|---|---|
| SimBA-DCT (Guo et al. 2019) | 353 | 100 |
| AdvFlow (Mohaghegh et al. 2020) | 841.4 | 100 |
| MetaAttack (Du et al. 2019) | 363.2 | 100 |
| CG-Attack (Feng et al. 2022) | 81.6 | 100 |
| EigenBA (Zhou et al. 2022) | 99 | 99.0 |
| **RLAB (ours)** | **60** | **100** |

Table 3: Evaluation of the proposed method with state-of-the-art on **ResNet-50** model trained on **CIFAR-10** dataset. Our approach achieved an average $L2$ of 1.3. Most of our competitors have not presented average L2 values.

Table 3 shows the performance of the proposed method against state-of-the-art attacks on the CIFAR-10 dataset. It can be observed that the success rate of our proposed method is the same as that of the competitors, which is 100 percent, while the average queries of the proposed approach outperform every state-of-the-art technique. Except for EigenBA (Zhou et al. 2022) and CG-Attack(Feng et al. 2022), which are close to our results, our approach beats the competitors by a large margin.

### Evaluating Different Filters

Table 4 represents the performance of the proposed approach with different filters. The table shows that the approach generates the best results for Gaussian noise, followed by the illumination filter for the ImageNet dataset. The best results were obtained for the brightness value of -0.1 with values ranging between (-1, 1). For DeadPixel, the percentage of pixels to be dropped for a given patch was set to 50 percent. For Gaussian blur, the standard deviation was set to 1. The standard deviation controls the amount of blurring with a larger value (> 1), creating significantly higher blurring compared to a smaller value. The performance of the Gaussian noise better than the other filters could be due to the strong nature of noise based distortion proved in numerous works (Neelakantan et al. 2015)(Poole, Sohl-Dickstein, and Ganguli 2014).

| Filter Type | AVG.Q | Max.$L_2$(avg.) | ASR % |
|---|---|---|---|
| **Gaussian Noise** | **166** | **4.74(2.48)** | 100 |
| Brightness | 266 | 6.94(3.93) | 100 |
| Gaussian Blur | 201 | 9.3(5.33) | 100 |
| Dead Pixel | 75 | 21.16(13.58) | 100 |

Table 4: Comparison of maximum L2, average queries and Average Success rate with different **distortion filters with RLAB**. (avg) represents average L2 over all samples **Dataset:** ImageNet, **Model:** ResNet-50

| Patch Size | AVG. Q | Average $L_2$ | ASR % |
|---|---|---|---|
| 2x2 | 178 | 2.48 | 100 |
| 4x4 | 197 | 11.29 | 100 |
| 8x8 | 188 | 17.52 | 100 |
| 16x16 | 133 | 32.16 | 100 |
| 32x32 | 114 | 63.45 | 100 |

Table 5: **Ablation study on different patch sizes** for Gaussian Noise filter. All the experiments were performed on the same set of images for a fair comparison. **Dataset:** ImageNet, **Model:** ResNet-50

### Performance vs. Complexity

In our proposed work, we generated all our results with the patch of size 2 × 2 and Noise distortion filter for best results. Even though the computation for sensitivity analysis primarily depends on the size of the image and can be accelerated, scaled, and batched on the GPU, there is a smaller variation based on the number of patches due to the post-processing overhead, which is typically around 10% of the total computation for 224 × 224 images with 2 × 2 patch sizes. Depending on the use case, our approach allows the use of different patch sizes at varying performance levels, represented in Table 5.

### Ablation Study on Filter Parameters

Table 6 shows the ablation study for hyperparameters and the highlighted values were chosen for the final results. Experiments showed that higher noise levels increased the final $L_2$ of the adversarial sample, while too little noise impacted the average number of queries. We evaluated the impact of different noise levels on the metrics as represented in table 6. We applied the same noise level for evaluation on both datasets (ImageNet, CIFAR-10) and all three victim models. We observed that the chosen noise level gave the best results across all datasets and victim models.

| Filter | AVG.Q | Avg. $L_2$ | ASR |
|---|---|---|---|
| Noise (var.) | | | |
| 0.0005 | 981 | 4.42 | 100 |
| 0.001 | 621 | 5.31 | 100 |
| **0.005** | **178** | **2.48** | **100** |
| 0.01 | 123 | 6.24 | 100 |
| Blur (std) | | | |
| **1** | **201** | **5.33** | **100** |
| 2 | 134 | 9.63 | 100 |
| brightness (intensity) | | | |
| **-0.1** | **266** | **6.94** | **100** |
| 0.1 | 241 | 7.21 | 100 |
| -0.5 | 82 | 8.51 | 100 |
| 0.5 | 90 | 8.97 | 100 |

Table 6: **Hyper-parameter tuning for different filters.** Table 4 represents the best configuration from this study. Experiments were performed with 1000 samples. **Dataset:** ImageNet, **Model:** ResNet-50

### Adversarial Retraining

We retrained the victim model with the adversarial samples for 5 additional epochs, with a reduced learning rate with adversarial samples generated from the training set. During every retraining procedure, the adversarial samples from the same original sample were generated for fairness. To evaluate the effectiveness of the adversarial retraining on CIFAR dataset, we used the CIFAR-10-C dataset as the benchmark which were constructed by corrupting the original CIFAR testsets. For each dataset, there are a total of 15 noise types with 5 level of intensities.

**Metrics** The clean error is the usual classification error on the clean or uncorrupted test data. As the corrupted test data appears at five different intensities $1 \leq s \leq 5$, and for a given corruption $c$, the error rate at corruption severity s is defined by $E_{c,s}$. Average error across these severities are calculated to generate unnormalized corruption error $uCE_c = \sum_{s=1}^{5} E_{c,s}$ and the average across all the corruptions is defined as the mean corruption Error. Robustness is not measured by how accurate the model predicts an outcome but to what extend the clean error is degraded when evaluated on a corrupted dataset. Hence, the Degradation of model performance is defined by,

$$mCE = \sum_{c=1} uCE_c \quad (5)$$

$$DegradationError = mCE - cleanError \quad (6)$$

We use the Degradation error to compare the performance of RLAB with other state-of-the-art competitors in this area. Furthermore, to measure the robustness of the retrained model with RLAB against other attacks, we use a standard metric, "Adversarial error". Adversarial error is defined as a model failure rate when the model is given adversarial examples to classify. The last column of table 9 shows the robustness of ResNet-50 retrained with RLAB's adversarial samples against two of the popular black-box attacks in the recent times.

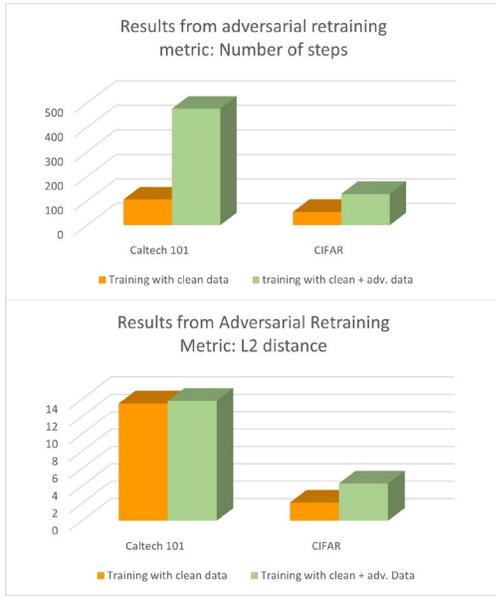

Figure 8: Results of Adversarial Retraining on CalTech-101 and CIFAR-10 datasets. (top) Steps define the number of steps taken for the model to miss-classify a given perturbed sample. (bottom) L2 is a metric that defines the deviation of the adversarial from the original sample.

|  |  | Classification Error (%) with Re-training | | |
|---|---|---|---|---|
| Dataset | Evaluated Against ↓ | SimBA Adv Training | Square Adv Training | **RLAB Adv Training** |
| CIFAR-10 | SimBA | - | 99.80 | **7.81** |
| CIFAR-10 | Square | 55.83 | - | **51.61** |
| CIFAR-10 | RLAB | 88.60 | 97.80 | - |
| Caltech-101 | SimBA | - | 2.15 | **1.37** |
| Caltech-101 | Square | 32.77 | - | **28.75** |
| Caltech-101 | RLAB | 75.00 | 75.04 | - |

Figure 9: Robustness comparison of our approach with Square and SimBA attack on ResNet-50 model with different datasets. Each attack was evaluated with the same 1000 samples generated from the testset.

### Results on Adversarial Retraining

Figure 8 shows the impact in the metrics, "Average steps" and "Average L2" after retraining for CIFAR-10 and Caltech-101 datasets. From 8(a), for both datasets, retraining with adversarial samples showed an increased number of steps to misclassify when compared to the base model. Similarly, from figure 8(b) retraining with adversarial samples demanded increased *l2* when compared to the base model.

Figure 9 shows retrained robustness of target model with RLAB is when compared to retraining with other two competitor approaches. The last column of the table shows the behavior of model retrained with adversarials amples from RLAB against other attacks to show the effect of improved robustness. Model retrained with RLAB when attacked with SimBA shows a low error of 7.81 % while a model retrained with SimBA and attacked with RLAB shows a high error rate of 88.6 %. Similarly, model retrained with RLAB when attacked with square has an error rate of 51.61 while the model retrained with square and attacked by RLAB shows an error rate of 97.80 which means that RLAB was able to foor 97 % of the times. This table shows that the model trained with RLAB can effectively defend themselves from other external attacks.

Table 7 shows comparison of RLAB's performance on CIFAR-10-C dataset with degradation error. Having a lower degradation error shows that the retrained model was not impacted even when evaluated on corrupted dataset (CIFAR-10-C) and a high degradation error shows that the training was very unstable and not robust enough. From the table, it can be observed that RLAB achieves a minimum degradation error when compared to the competitors.

Table 7: Degradation error% for Image classification architectures on CIFAR-10-C for state-of-the-art techniques. For fairness, all of the techniques were evaluated with the same seed.

| Model | Mixup (?) | Cutmix (?) | Augmix (?) | RLAB |
|---|---|---|---|---|
| ResNet-50 | 29.0 | 31.5 | 13 | **6.0** |
| DenseNet | 24.0 | 33.5 | 15 | **11** |
| Inception-V3 | 29 | 23 | 11.5 | **9.5** |
| Mean | 27.3 | 29.3 | 13.1 | 8.83 |

This paper is inspired by earlier work (Sarkar et al. 2024a,b, 2023b,a,c, 2022).

## Extensions of RLAB

**Visual Explanation:** RLAB can also be extended to visual explanations for image classification models.

**Signals and Video:** RLAB is effective as an adversarial attack tool to address robustness beyond image classifiers to models classifying signals, 3D images, and multi-modal satellite images.

## Conclusion

The use of RL strengthens the RLAB platform, which can optimize black-box adversarial attacks with different types of distortion filters or a mix of them. With the "Bring Your Own Filters - BYOF" approach, the RLAB platform supports any new types of distortion relevant to specific real-life use cases. This also helps assess the robustness against the most pertinent non-malicious adversarial perturbations. Compared to the hand-crafted heuristics of most state-of-the-art adversarial attacks, this approach expands the scope and applicability to new distortion filters. As part of future work, we are expanding the scope of RLAB for adversarial attack-based evaluation of other elements of trustworthiness like bias and fairness of both computer vision applications and natural language processing.